\documentclass{article} % For LaTeX2e
\usepackage{iclr2026_conference,times}

\usepackage{amsmath,amsfonts,bm}

\def\eqref#1{equation~\ref{#1}}
\def\1{\bm{1}}
\DeclareMathAlphabet{\mathsfit}{\encodingdefault}{\sfdefault}{m}{sl}
\SetMathAlphabet{\mathsfit}{bold}{\encodingdefault}{\sfdefault}{bx}{n}

\usepackage[utf8]{inputenc} % allow utf-8 input
\usepackage[T1]{fontenc}    % use 8-bit T1 fonts
\usepackage{url}            % simple URL typesetting
\usepackage{booktabs}       % professional-quality tables
\usepackage{amsfonts}       % blackboard math symbols
\usepackage{nicefrac}       % compact symbols for 1/2, etc.
\usepackage{microtype}      % microtypography
\usepackage{xcolor}         % colors
\usepackage{graphicx}
\usepackage{enumitem}
\usepackage{xspace}
\usepackage{caption}
\usepackage{ulem}
\usepackage[pagebackref=true,breaklinks=true,colorlinks,bookmarks=false]{hyperref}

\def\eg{\textit{e.g.}}
\def\ie{\textit{i.e.}}

\newcommand{\methodname}{4DNeX\xspace}
\newcommand{\datasetname}{4DNeX-10M\xspace}

\title{\methodname: Feed-Forward 4D Generative Modeling Made Easy}

\author{Zhaoxi Chen$^{1^{\ast}}$, 
Tianqi Liu$^{1^{\ast}}$, 
Long Zhuo$^{2^{\ast}}$,
Jiawei Ren$^{1}$, 
Zeng Tao$^{2}$, 
He Zhu$^{2}$, \\
\textbf{Qifeng Liang$^{3}$,
Hanfei Zhao$^{4}$,
Fangzhou Hong$^{1}$, 
Liang Pan$^{2^{\dagger}}$, 
Ziwei Liu$^{1^{\dagger}}$}\\
$^{1}$S-Lab, Nanyang Technological University \quad
$^{2}$Shanghai AI Laboratory \\
$^{3}$University of California, San Diego \quad
$^{4}$Duke University
}

\iclrfinalcopy % Uncomment for camera-ready version, but NOT for submission.
\newcommand\nnfootnote[1]{%
  \begin{NoHyper}
  \renewcommand\thefootnote{}\footnote{#1}%
  \addtocounter{footnote}{-1}%
  \end{NoHyper}
}

\begin{document}

\maketitle

\nnfootnote{$\ast$ Equal contribution. $\dagger$ Corresponding authors.}

\begin{center}
    \vspace{-47pt}
    \textbf{\url{https://4dnex.github.io/}}
    \vspace{5pt}
    \centering
    \captionsetup{type=figure}
    \includegraphics[width=\textwidth]{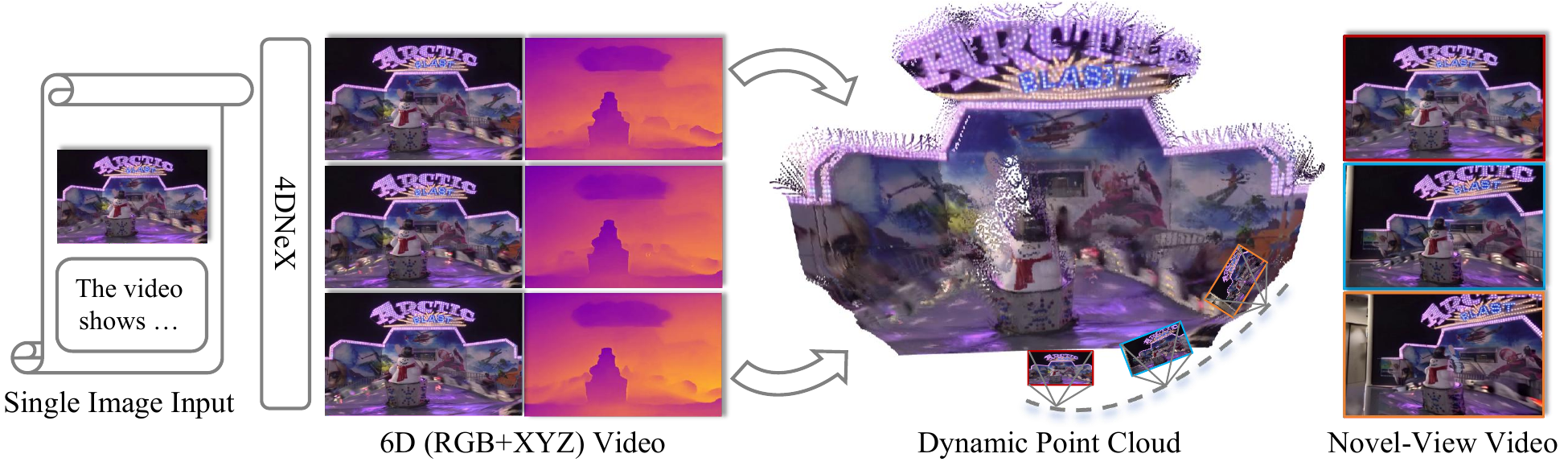}
    \caption{\textbf{\methodname generates 6D video from a single image to enable 4D scene creation and novel-view video rendering.}}
    \label{fig:teaser}
\end{center}%

\begin{abstract}

We present \textbf{\methodname}, the first feed-forward framework for generating 4D (\ie, dynamic 3D) scene representations from a single image. In contrast to existing methods that rely on computationally intensive optimization or require multi-frame video inputs, \methodname enables efficient, end-to-end image-to-4D generation by fine-tuning a pretrained video diffusion model.
Specifically, 
\textbf{1)} to alleviate the scarcity of 4D data, we construct \datasetname, a large-scale dataset with high-quality 4D annotations generated using advanced reconstruction approches. 
\textbf{2)} we introduce a unified 6D video representation that jointly models RGB and XYZ sequences, facilitating structured learning of both appearance and geometry.
\textbf{3)} we propose a set of simple yet effective adaptation strategies to repurpose pretrained video diffusion models for 4D modeling.
\methodname produces high-quality dynamic point clouds that enable novel-view video synthesis. Extensive experiments demonstrate that \methodname outperforms existing 4D generation methods in efficiency and generalizability, offering a scalable solution for image-to-4D modeling and laying the foundation for generative 4D world models that simulate dynamic scene evolution.

\end{abstract}
\section{Introduction}
\label{sec:intro}

The images we capture are 2D projections of the 4D (\ie, dynamic 3D) physical world. Creating a 4D scene from such 2D observations, particularly from a single image, is a highly challenging yet compelling task.
As a core capability in generative modeling, image-to-4D generation lays the foundation for building 4D world models that can predict and simulate dynamic scene evolution, enabling a wide range of applications in AR/VR, film production, and digital content creation. 
% with promising applications in AR/VR, film production, and digital content creation.

Existing approaches for 4D scene modeling can be broadly classified into two categories. The first comprises 4D generation methods, which typically adopt representations such as Neural Radiance Fields (NeRF)~\cite{nerf} or 3D Gaussian Splatting (3DGS)~\cite{3dgs}. These methods can be further divided into feed-forward~\cite{l4gm,cat4d,genXD,dimensionX} and optimization-based variants~\cite{free4d,4real,dream-in-4d,4dfy,animate124,dreamgaussian4d}. However, they either require video input or rely on object-centric, computationally intensive optimization procedures. The second category includes dynamic Structure-from-Motion (SfM) approaches~\cite{li2024megasam,monst3r,geometrycrafter,geo4d,cut3r}, which estimate dynamic 3D structures such as time-varying point clouds from video sequences. However, these methods remain incapable of generating 4D representations from a single image.

To this end, we aim to develop a feed-forward framework for 4D scene generation from a single image. A straightforward solution is to fine-tune a pretrained video diffusion model. However, this approach presents two core challenges: \textbf{1)} how to mitigate the scarcity of 4D data, and \textbf{2)} how to adapt the pretrained model in a simple and efficient way.

For the \textbf{first} challenge, we curate \datasetname, a large-scale dataset comprising both static and dynamic scenes, with high-quality 4D annotations generated from monocular videos using state-of-the-art reconstruction methods~\cite{dust3r,vggt,pi3,monst3r,li2024megasam}. To ensure geometric accuracy and scene diversity, we apply careful data selection, pseudo-annotation generation, and multi-stage filtering.
To address the \textbf{second} challenge, we first introduce a unified 6D video representation that models RGB and XYZ sequences jointly, enabling the structured modeling of both appearance and geometry. We then systematically investigate different fusion strategies between the two modalities and show that width-wise fusion achieves the most effective cross-modal alignment. Moreover, we incorporate a set of carefully designed techniques, including XYZ initialization, XYZ normalization, mask design, and modality-aware token encoding, to adapt pretrained video diffusion models in a simple manner while preserving their generative priors.

To summarize, we present \methodname, the first feed-forward framework for image-to-4D generation (Fig.~\ref{fig:teaser}). We qualitatively demonstrate the plausibility of the generated dynamic point clouds. Furthermore, to validate their utility, we leverage TrajectoryCrafter~\cite{trajectorycrafter} to transform the generated 4D point clouds into novel-view videos, achieving comparable results to existing 4D generation methods. In addition, we perform comprehensive ablation studies to validate the effectiveness of our proposed fine-tuning strategies.

Our main contributions can be summarized as follows:
\begin{itemize}[leftmargin=*]
    \item We propose \methodname, the first feed-forward framework for image-to-4D generation, capable of producing dynamic point clouds from a single image.
    \item We construct \datasetname, a large-scale dataset with high-quality 4D annotations.
    \item We introduce a set of simple yet effective fine-tuning strategies to adapt pretrained video diffusion models for 4D generation.
\end{itemize}

\section{Related Work}
\label{sec:review}

\subsection{Optimization-based 4D Generation}
Recent work has explored optimization-based methods for 4D generation. Leveraging the priors of pre-trained diffusion models~\cite{cfg,ddim,ddpm}, they optimize 3D and 4D representations~\cite{nerf,d-nerf,3dgs,4dgs} using the synthesized dynamic multi-view images or score distillation sampling~\cite{dreamfusion}. A core challenge for these approaches lies in ensuring the temporal and spatial consistency of the acquired guidance. Some studies~\cite{4real,DBLP:journals/corr/abs-2412-20422,dream-in-4d,animate124,gaussianflow, 4dfy,consistent4d,stag4d} build upon 3D representations optimized from static images and incorporate dynamic information derived from video diffusion models to refine 3D into 4D. Other works~\cite{eg4d,pan2024efficient4d,4dgen,make-a-video4d,dreamgaussian4d,free4d} initiate from video generation, aiming for cross-view consistency to facilitate the optimization of 4D representations. The most recent method, Free4D~\cite{free4d}, first generates multi-view videos in a training-free manner through a set of consistency-preserving designs, and then optimizes a 4D representation. However, it is limited to relatively small camera and scene motion. In addition to the inherent challenge of maintaining consistent guidance, optimization-based methods also suffer from high computational cost, long runtime, and instability caused by multi-stage optimization.In this work, we propose a feed-forward 4D generation framework that produces 4D representations, offering a more efficient and scalable alternative.

\subsection{Feed-forward 4D Generation}

Feed-forward 4D generation methods aim to directly predict 4D representations from input via a single forward pass, avoiding the computational cost and inconsistency of optimization-based pipelines. This enables efficient, end-to-end learning of spatiotemporal structures. Some works~\cite{genXD,dimensionX,gen3c,deepmind_genie3_2025} focus on generating temporally consistent and viewpoint-controllable videos. For example, GenXD~\cite{genXD} concatenates camera and image conditions and employs multi-view-temporal fusion modules, but still requires post-optimization to obtain explicit 4D geometry. DimensionX~\cite{dimensionX} uses motion-specific LoRA modules for dynamic view synthesis, but lacks support for fully free-view generation.
Other methods aim to directly generate 4D representations. L4GM~\cite{l4gm} extends LGM~\cite{lgm} by predicting per-frame 3D Gaussian splats and using temporal self-attention to ensure consistency. Cat4D~\cite{cat4d} finetunes CAT3D~\cite{cat3d} on pseudo-4D data, but may struggles to generalize beyond specific video generation sources.
TesserAct~\cite{zhen2025tesseract} tackles 4D prediction in embodied robotics settings by jointly predicting RGB, depth, normals, and motion from a single image. However, it targets task-specific representations (\eg, surface normals), relies on heavy multitask learning, and is not designed for general, in-the-wild scenarios.
In contrast, we aim to efficiently generate general-purpose 4D representations from a single image by leveraging pre-trained video diffusion models and introducing a transferable training paradigm.
Another research line includes dynamic Structure-from-Motion (SfM) approaches~\cite{li2024megasam,monst3r,geometrycrafter,geo4d,cut3r,vipe}, which recover time-varying 3D structures, such as dynamic point clouds, from multi-frame videos. However, these methods cannot generate 4D representations from a single image. Fundamentally, they focus on reconstructing dynamic geometry from dense video input, while we tackle the more challenging task of jointly generating appearance and geometry sequences from a single image.

\subsection{Video Generation Model}

Pre-trained video generation models~\cite{VDM,DBLP:journals/corr/abs-2210-02303,kling,make-a-video} have demonstrated remarkable capabilities and underpin numerous downstream tasks. CogVideo~\cite{cogvideo} and CogVideoX~\cite{cogvideox} employ specifically designed expert transformers and 3D full attention mechanisms to achieve high-quality text-to-video generation. Building upon text-to-video synthesis, DynamiCrafter~\cite{dynamicrafter} enables the animation of input images at arbitrary positions within a video.
Beyond classical video generation tasks, significant efforts have focused on generating videos from target viewpoints. SynCamMaster~\cite{SynCamMaster} and Collaborative Video Diffusion~\cite{Collaborative-Video-Diffusion} encode camera viewpoints and leverage multi-view synchronization to generate paired videos based on text-to-video models.
Furthermore, several works aim to integrate the capabilities of video generation models into the 3D domain, particularly for post-processing multi-view reconstruction results. ViewCrafter~\cite{viewcrafter} introduces video generation models to the 3D domain to refine lossy reconstructions from different viewpoints, yielding complete novel view images. TrajectoryCrafter~\cite{trajectorycrafter} introduces a data construction paradigm for handling novel view synthesis of dynamic scenes. Our approach utilizes TrajectoryCrafter~\cite{trajectorycrafter} to process the generated 4D point clouds, transforming them into novel videos with target viewpoints.

\section{\datasetname}
\label{sec:dataset}

\begin{figure}
    \centering
    \includegraphics[width=1\linewidth]{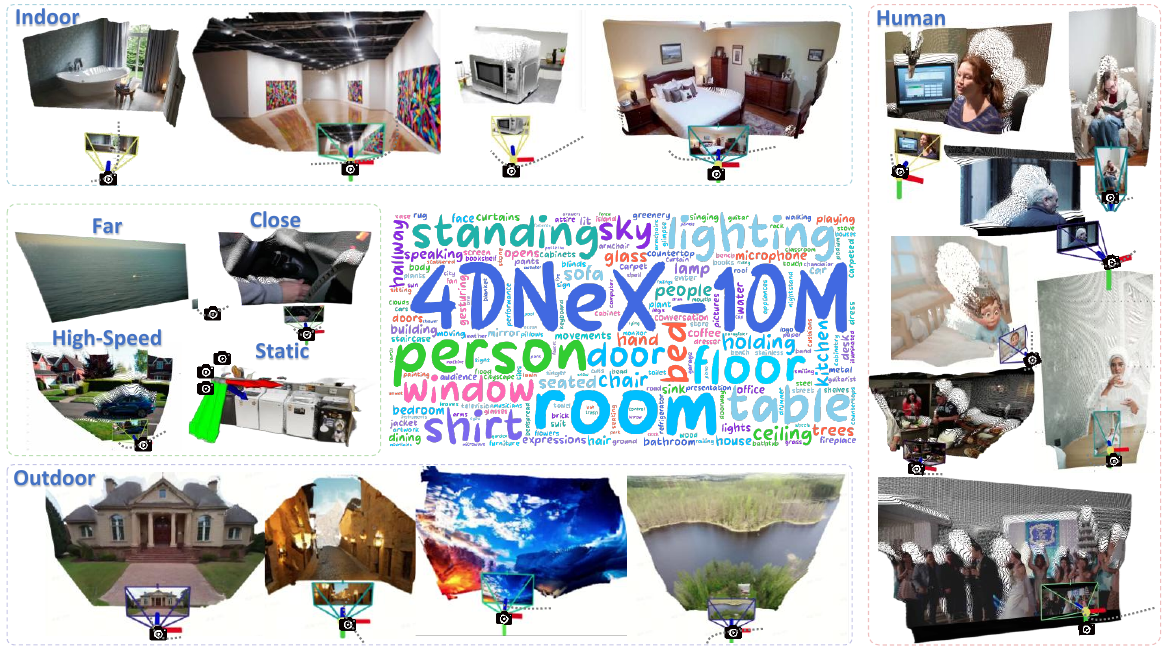}
    \caption{\textbf{Visualization of \datasetname Dataset.}
    Our dataset spans a wide range of dynamic scenarios, including indoor, outdoor, close-range, far-range, static, high-speed, and human-centric scenes. The word cloud summarizes common visual concepts captured in the dataset, while the 4D point clouds and camera trajectories demonstrate the spatial precision of our pseudo-annotations.}
    \label{fig:data_vis}
\end{figure}

\begin{figure}
    \centering
    \includegraphics[width=1\linewidth]{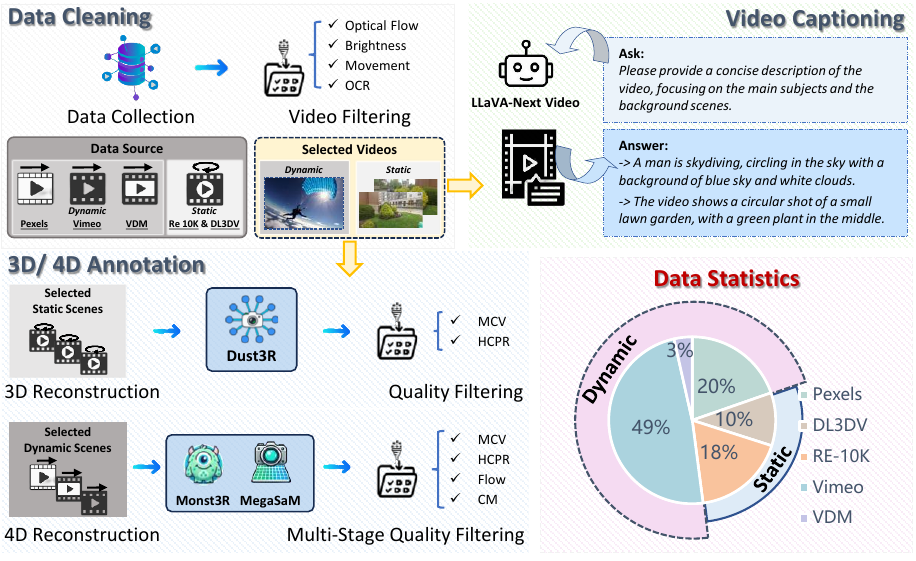}
    % \vspace{-7mm}
     \caption{\textbf{Data Curation Pipeline.} The video data is collected from various sources and then selected by video filtering during Data Cleaning. The selected data is captioned via LLaVA-Next-Video model in Video Captioning. The selected data is processed and finally filtered out the video with high-quality annotation during 3D/4D Annotation. Data statistics is also provided in bottom right.}
    \label{fig:data_pipe}
\end{figure}

To address the data scarcity in 4D generative modeling, we introduce \datasetname, a large-scale hybrid dataset tailored for training feed-forward 4D generative models. It aggregates videos from public sources and internal pipelines, encompassing both static and dynamic scenes. All data undergoes rigorous filtering, pseudo-annotation, and quality assessment to ensure geometric consistency, motion diversity, and visual realism. As shown in Figure~\ref{fig:data_vis}, our proposed dataset encompasses a highly diverse range of scenes, including indoor and outdoor environments, distant landscapes, close-range settings, high-speed scenarios, static scenes, and human-inclusive situations. Furthermore, \datasetname encompasses a wide variety of lighting conditions and a profusion of human activities. Meanwhile, we provide precise 4D pointmaps and camera trajectories of these corresponding scenes. In total, \datasetname contains over 9.2 million video frames with pseudo annotations. 
For data curation, as illustrated in Figure~\ref{fig:data_pipe}, we curate this data using an automated acquisition and filtering pipeline comprising several stages: 1) data cleaning, 2) data captioning, and 3) 3D/4D annotation.

\subsection{Data Preprocessing}

The foundation of \datasetname is built upon a variety of datasets, each contributing distinct scene characteristics and motion types.

\textbf{Data Sources.}
We collect monocular videos from several sources. DL3DV-10K (DL3DV)~\cite{dl3dv} and RealEstate10K (RE10K)~\cite{realestate10k} offer static indoor and outdoor videos with diverse camera trajectories. The Pexels dataset provides a large pool of human-centric stock videos with auxiliary metadata such as movements, OCR, and optical flow. The Vimeo Dataset, selected from Vchitect 2.0~\cite{fan2025vchitect}, contributes in-the-wild dynamic scenes. Synthetic data sourced from Vbench~\cite{vbench} contains dynamic sequences using video diffusion models (VDM).

\textbf{Initial Filtering.}
For large-scale sources like Pexels, we apply metadata filtering, including optical flow, motion, and OCR, to eliminate non-compliant videos, such as those exhibiting excessive motion blur or text-saturated videos. Across all data sources, brightness filtering is applied based on average luminance ($0.299R+0.587G+0.114B$) to discard videos with extreme illumination conditions.

\textbf{Video Captioning.}
For datasets without textual annotations (\eg, DL3DV-10K and RE-10K), we use LLaVA-Next-Video~\cite{llavanextvideo} to generate captions. We sample 32 frames uniformly per video (or clip) and feed them to the LLaVA-NeXT-Video-7B-Qwen2 model with the prompt: \textit{"Please provide a concise description of the video, focusing on the main subjects and the background scenes."} For scenes with consistent content (\eg, DL3DV-10K, Dynamic Replica), we generate one caption per video. For RealEstate10K, we split each video into clips and caption them separately.

\subsection{Static Data Processing}

% \textbf{Raw Static Set.}
To learn strong geometric priors, we curate static monocular videos from DL3DV-10K~\cite{dl3dv} and RE-10K~\cite{realestate10k}. These cover a wide range of environments including homes, streets, stores, and landmarks, with varied camera trajectories providing rich multi-view coverage.

\textbf{Pseudo 3D Annotation.}
As these datasets lack 3D ground-truth, we employ DUSt3R~\cite{dust3r}, a stereo reconstruction model, to generate pseudo point maps. For each video, DUSt3R is applied exhaustively over view pairs to form a view graph, followed by global fusion (per the original paper) to recover a consistent scene-level 3D structure.

\textbf{Quality Filtering.}
To ensure high-quality annotations, we define two metrics using the confidence maps from DUSt3R: 1) the \textit{Mean Confidence Value (MCV)}, averaging pixel-wise confidence scores over all frames, and 2) the \textit{High-Confidence Pixel Ratio (HCPR)}, representing the proportion of pixels exceeding a threshold $\tau$. We select the top-$r\%$ of clips for each metric and retain over 100K high-quality 28-frame clips with reliable pseudo point map annotations for static training.

\subsection{Dynamic Data Processing}

% We further enrich \datasetname\ with dynamic videos sourced from Pexels, VDM, and Vimeo. For clips without ground-truth geometry, we employ MonST3R~\cite{monst3r} and MegaSaM~\cite{li2024megasam} to reconstruct time-varying point maps and camera trajectories.
To enrich \datasetname with dynamic content, we collect monocular videos from Pexels, VDM, and Vimeo. These datasets contain diverse real-world scenes with motion and depth variation but lack ground-truth geometry.

\textbf{Pseudo 4D Annotation.}
% MonST3R and MegaSaM are advanced dynamic reconstruction models that produce temporally coherent 3D point clouds and globally aligned camera poses from monocular videos, enabling the generation of pseudo 4D annotations for real-world dynamic scenes.
We employ MonST3R~\cite{monst3r} and MegaSaM~\cite{li2024megasam}, two advanced dynamic reconstruction models, to generate pseudo 4D annotations. Each model recovers temporally coherent 3D point clouds and globally aligned camera poses from monocular videos, enabling the construction of time-varying scene representations.

\textbf{Multi-Stage Filtering.}
To select high-quality clips, we apply three sequential filtering strategies. First, we use the final alignment loss in the global fusion stage, which reflects multi-view consistency and flow agreement with RAFT~\cite{raft}, to filter out low-quality reconstructions. Second, we assess camera smoothness (CS) by computing frame-wise velocity and acceleration from camera translations, and estimate local trajectory curvature as:
\begin{equation}
\kappa_i = \frac{\|\mathbf{v}_{i+1} - \mathbf{v}_i\|}{\|\mathbf{v}_{i+1}\|^2 + \|\mathbf{v}_i\|^2 + \epsilon}, \quad \epsilon > 0.
\end{equation}
Clips with low average velocity, acceleration, and curvature are retained. Third, we apply the same \textit{Mean Confidence Value (MCV)} and \textit{High-Confidence Pixel Ratio (HCPR)} used in the static pipeline.
After filtering, we retain approximately 32K clips from the MonST3R-processed set, 5K clips from VDM, and 27K from Pexels, and over 80k clips from MegaSaM-processed set. Together, these yield a total over 110K high-quality clips with pseudo 4D annotations, enabling robust modeling of dynamic 3D scenes across a wide range of motions and appearances.

\section{\methodname}
\label{sec:method}

\begin{figure}[t]
    \centering
    \includegraphics[width=0.95\textwidth]{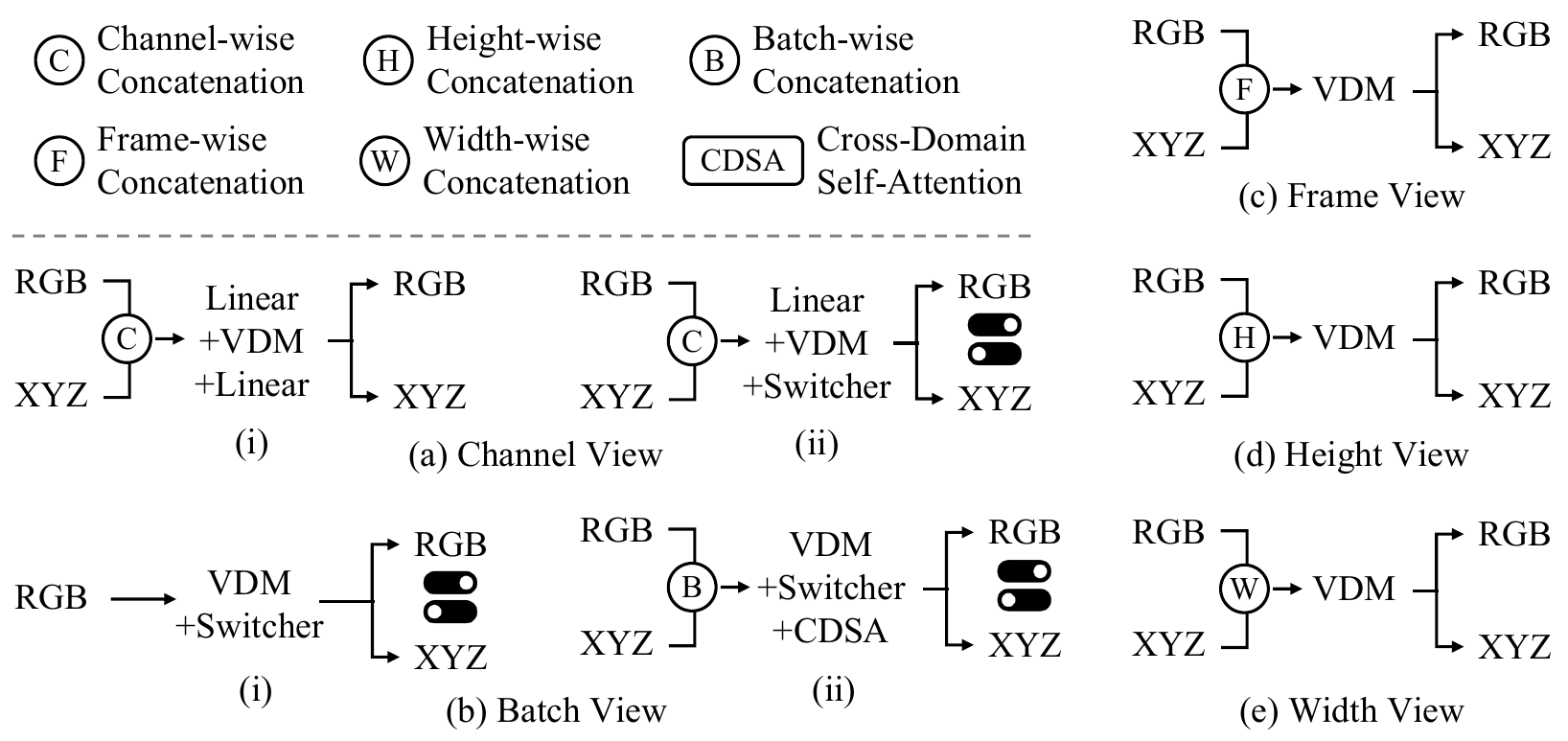}
    \caption{\textbf{Comparison of fusion strategies for joint RGB and XYZ modeling.} We explore five fusion strategies and analyze their impact on model compatibility and cross-modal alignment.}
    \label{fig:fusion}
\end{figure}

\subsection{Problem Formulation}
\label{subsec:formulation}
Given a single image $I_0 \in \mathbb{R}^{H \times W \times 3}$, we aim to construct a 4D (\ie, dynamic 3D) representation of the underlying scene geometry. This task can be formulated as learning a conditional distribution over a sequence of dynamic point clouds:
\begin{equation}
p\left(\{P_t\}_{t=0}^{T-1} \mid I_0\right),
\end{equation}
where \( \{P_t\}_{t=0}^{T-1} \) denotes the sequence of dynamic point clouds. However, directly modeling point clouds is challenging due to their highly unstructured nature.
To address this, inspired by~\cite{zhang2025world}, we adopt a pixel-aligned point map representation, \textit{XYZ}, where each frame $X_t^{\text{XYZ}} \in \mathbb{R}^{H \times W \times 3}$ encodes the 3D coordinates of each pixel in the global coordinates. This format provides a structured and learnable structure, making it compatible with existing generative models. Instead of directly modeling \( \{P_t\} \), we reformulate the problem as predicting paired RGB and XYZ image sequences:
\begin{equation}
p\left(\{X^{RGB}_t,\ X^{XYZ}_t\}_{t=0}^{T-1} \mid I_0\right).
\end{equation}
Accordingly, the joint distribution can be also factorized as:
\begin{equation}
p\left(\{X^{RGB}_t\}_{t=0}^{T-1},\ \{X^{XYZ}_t\}_{t=0}^{T-1} \mid I_0\right).
\end{equation}
Therefore, a 4D scene can be effectively represented using a 6D video composed of paired RGB and XYZ sequences.
This simple and unified representation offers two key advantages: it enables explicit 3D consistency supervision through pixel-aligned XYZ maps, and eliminates the need for camera control, facilitating scalable and robust 4D generation.

To model this distribution, we adopt Wan2.1~\cite{wan2025wan}, a video diffusion model trained under the flow matching~\cite{flow_matching} framework. We extend its image-to-video capability to generate 6D videos as \( V = \{X^{RGB}_t,\ X^{XYZ}_t\}_{t=0}^{T-1} \).
\( V \) is first encoded into a latent space via a VAE encoder \( \mathcal{E} \):
$x_1 = \mathcal{E}(V)$, and interpolating with a noise latent $x_0 \sim \mathcal{N}(0, I)$:
\begin{equation}
x_t = (1 - t)x_0 + t x_1, \quad t \sim \mathcal{U}(0, 1).
\end{equation}
And a velocity predictor \( u \) is trained to regress the velocity between endpoints:
\begin{equation}
\mathcal{L}_{\text{FM}} = \mathbb{E} \left[ \left\| u(x_t, c_{\text{img}}, c_{\text{txt}}, t) - (x_1 - x_0) \right\|^2 \right],
\end{equation}
where \( c_{\text{img}} \) and \( c_{\text{txt}} \) denote the image and text condition embeddings.
This formulation enables efficient learning of temporally coherent and geometrically consistent 6D video sequences.

\subsection{Fusion Strategies}
\label{subsec:fusion}

To finetune the video diffusion model for joint RGB and XYZ generation, a key challenge is designing an effective fusion strategy that enables the model to leverage both modalities. Our goal is to exploit the strong priors of pretrained models through simple yet effective fusion designs.
Motivated by prior work, latent concatenation is a widely adopted technique for joint modeling. We systematically explore fusion strategies across different dimensions, as illustrated in Fig.~\ref{fig:fusion}.

\textbf{Channel-wise Fusion.} A straightforward approach is to concatenate RGB and XYZ along the channel dimension, and insert a linear layer (\textit{a.i}) or a modality switcher (\textit{a.ii}) to adapt the input and output formats. However, this strategy disrupts the input and output distributions expected by the pretrained model, which undermines the benefits of pretraining. It typically requires large-scale data and substantial computational resources to achieve satisfactory performance.

\textbf{Batch-wise Fusion.} To maintain pretrained distributions, this strategy treats RGB and XYZ as separate samples and uses a switcher to control the output modality (\textit{b.i}). While it preserves unimodal performance, it fails to establish cross-modal alignment. Even with additional cross-domain attention layers (\textit{b.ii}), the modalities remain poorly correlated.

\begin{figure}[t]
\centering
\includegraphics[width=\textwidth]{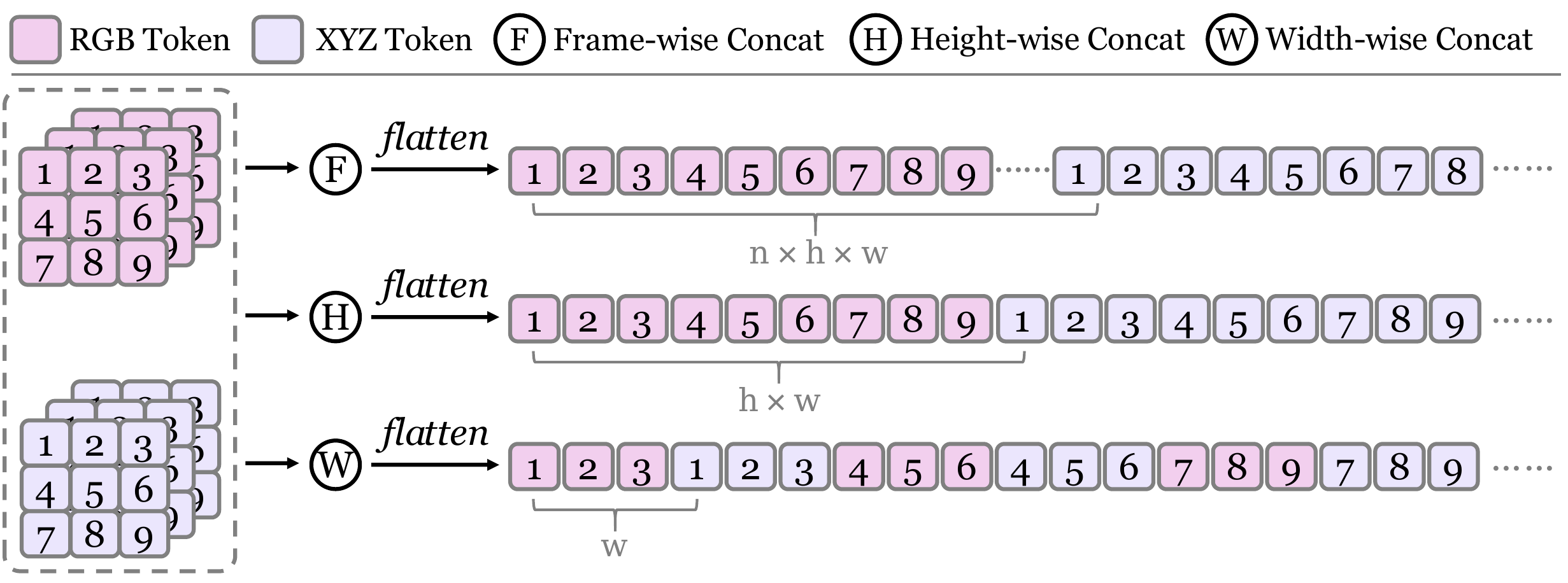}
\caption{\textbf{Comparison of spatial fusion strategies.} 
We compare frame-, height-, and width-wise fusion in terms of the interaction distance between RGB and XYZ tokens.}
\label{fig:analysis}
\end{figure}

\textbf{Frame-/Height-/Width-wise Fusion.}  
These strategies concatenate RGB and XYZ along the frame~(\textit{c}), height~(\textit{d}), or width~(\textit{e}) dimensions, preserving the distributions of the pretrained model while enabling cross-modal interaction within a single sample. We analyze them from the perspective of token interaction distance. Intuitively, shorter interaction distance between corresponding tokens makes it easier for the model to learn cross-modal alignment. As shown in Fig.~\ref{fig:analysis}, width-wise fusion yields the shortest interaction distance, leading to more effective alignment and higher generation quality, as confirmed by our experiments (Sec.~\ref{subsec:ab}).

\subsection{Network Architecture}
\label{subsec:network}

\begin{figure}[t]
\centering
\includegraphics[width=\textwidth]{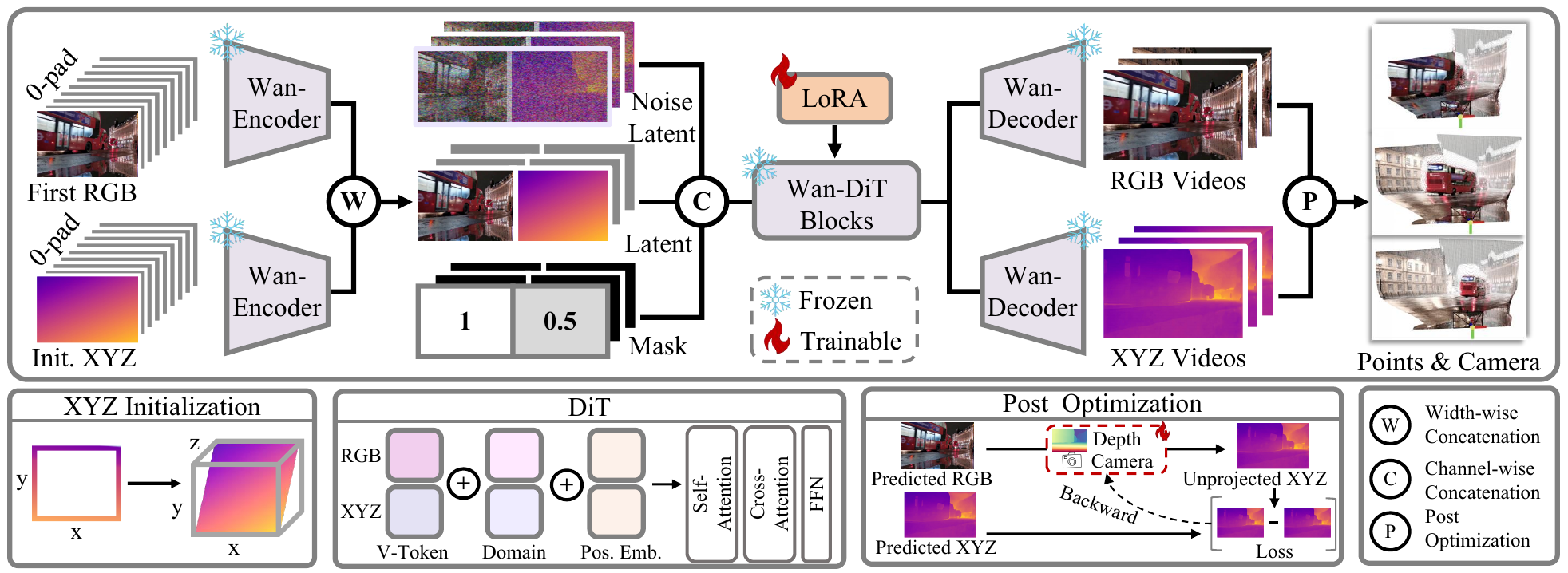}
\caption{\textbf{Overview of \methodname.}
Given a single RGB image and an initialized XYZ map, \methodname encodes both inputs with a VAE encoder and fuses them via width-wise concatenation. The fused latent, combined with a noise latent and a guided mask, is processed by a LoRA-tuned Wan-DiT model to jointly generate RGB and XYZ videos. A lightweight post-optimization step recovers camera parameters and depth maps from the predicted outputs.
}
\label{fig:pipeline}
\end{figure}

As illustrated in Fig.~\ref{fig:pipeline}, our framework takes a single image \( I_0 \in \mathbb{R}^{H \times W \times 3} \) and an initialized XYZ map \( X^{init} \in \mathbb{R}^{H \times W \times 3} \) as conditions. Both are encoded by a frozen VAE encoder and concatenated along the width dimension. This fused condition is then combined with a noise latent \( x_t \) and a binary mask \( M \) along the channel dimension, and fed into a pretrained DiT with LoRA tuning. The output latent is decoded by a VAE decoder to generate paired RGB and XYZ video sequences. A lightweight post-optimization step further recovers camera parameters and depth maps from the predicted outputs.

\textbf{XYZ Initialization.}  
We initialize the first-frame XYZ map \( X^{init} \) using a sloped depth plane. Specifically, we define a normalized 2D coordinate grid over the range \([-1, 1]^2\) and compute the initial XYZ values as:
\begin{equation}
X^{init}_{i,j} = \left( \frac{2j}{W - 1} - 1,\ \frac{2i}{H - 1} - 1,\ \frac{2i}{H - 1} - 1 \right).
\end{equation}
This results in a sloped plane where depth values gradually increase from the bottom to the top of the image, reflecting common depth priors in natural scenes (\eg, sky regions appearing farther away). Such initialization provides a stable starting point for geometry learning.

\textbf{XYZ Normalization.}
Since the VAE is pretrained on RGB images, directly encoding XYZ inputs with different distributions can cause instability and suboptimal performance. To mitigate this issue, inspired by~\cite{3dtopiaxl} , we apply a modality-aware normalization strategy to adapt the XYZ latent to the pretrained VAE's distributional priors. Specifically, we compute the mean \(\mu\) and standard deviation \(\sigma\) of XYZ latent across the training dataset, and normalize the encoded representation as:
\begin{equation}
\hat{x} = \frac{x - \mu}{\sigma},
\end{equation}
where \(x\) denotes the XYZ latent. Before passing into the VAE decoder, we perform de-normalization to recover the original scale:
\begin{equation}
x = \hat{x} \cdot \sigma + \mu.
\end{equation}

\textbf{Mask Design.}
Following~\cite{wan2025wan}, we introduce a guided mask \( M \in [0,1]^{T \times H \times W} \), where \( M_{t,i,j} = 1 \) indicates a known pixel and \( M_{t,i,j} = 0 \) indicates a pixel to be generated. Since we use an approximate initialization for the first-frame XYZ map, we assign a soft mask:
\begin{equation}
M^{XYZ}_{0,i,j} = 0.5, \quad \forall\, i,j,
\end{equation}
which encourages the model to refine the initial geometry during generation.

\textbf{Modality-Aware Token Encoding.}
To preserve pixel-wise alignment across modalities during joint modeling, we adopt a shared rotary positional encoding (RoPE)~\cite{rope} for RGB and XYZ tokens.  
To further distinguish their semantic differences, we introduce a learnable domain embedding.  
Given RGB and XYZ token sequences \( x^{\text{RGB}}, x^{\text{XYZ}} \in \mathbb{R}^{L \times D} \), we apply the following encoding:
\begin{equation}
\begin{aligned}
x^{RGB} &\leftarrow \mathrm{RoPE}(x^{RGB}) + e_{RGB}, \\
x^{XYZ} &\leftarrow \mathrm{RoPE}(x^{XYZ}) + e_{XYZ},
\end{aligned}
\end{equation}
where \( \mathrm{RoPE}(\cdot) \) denotes the shared rotary positional encoding, and \(e_{RGB}, e_{XYZ} \in \mathbb{R}^{1 \times D} \) are learnable domain embeddings broadcasted across the sequence.

\textbf{Post-Optimization.}
Since our method produces XYZ videos that represent dense 3D points in global coordinates, we can recover the corresponding camera parameters \( C = (R, t, K) \) and depth maps \( d \) for the generated RGB frames via a lightweight post-optimization step. Specifically, we minimize the reprojection error between the generated and back-projected 3D coordinates:
\begin{equation}
\min_{R, t, K, d} \sum_{i,j} \left\| \tilde{q}^{XYZ}_{i,j} - \hat{q}^{XYZ}_{i,j} \right\|_2^2,
\end{equation}
where \( \hat{q}^{XYZ}_{i,j} \) denotes the generated 3D coordinate, and \( \tilde{q}^{XYZ}_{i,j} \) is computed by back-projecting the depth value into 3D space:
\begin{equation}
\tilde{q}^{XYZ}_{i,j} = [R \mid t]^{-1} K^{-1} \left( d_{i,j} \cdot [i, j, 1]^\top \right).
\end{equation}
This optimization is computationally efficient and can be parallelized across views, producing physically plausible and geometrically consistent estimates of camera poses and depth maps.

\section{Experiments}
\label{sec:exp}

\begin{figure}[t]
\centering
\includegraphics[width=\textwidth]{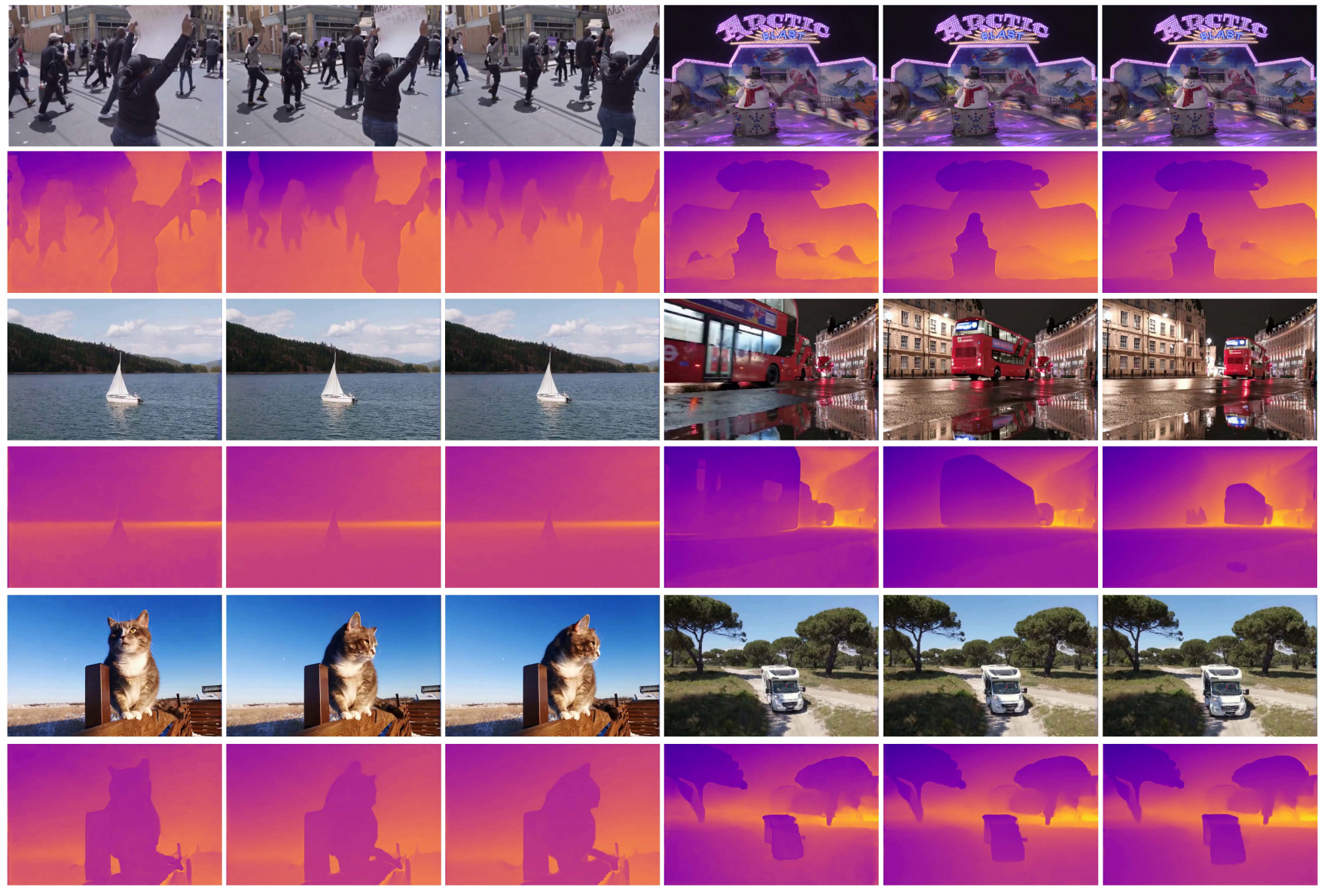}
\caption{\textbf{Generated RGB and XYZ sequences from single-image input.} 
Each pair of rows shows the output RGB video and its corresponding XYZ sequence.}
\label{fig:geo_result}
\end{figure}

\begin{figure}[t]
\centering
\includegraphics[width=\textwidth]{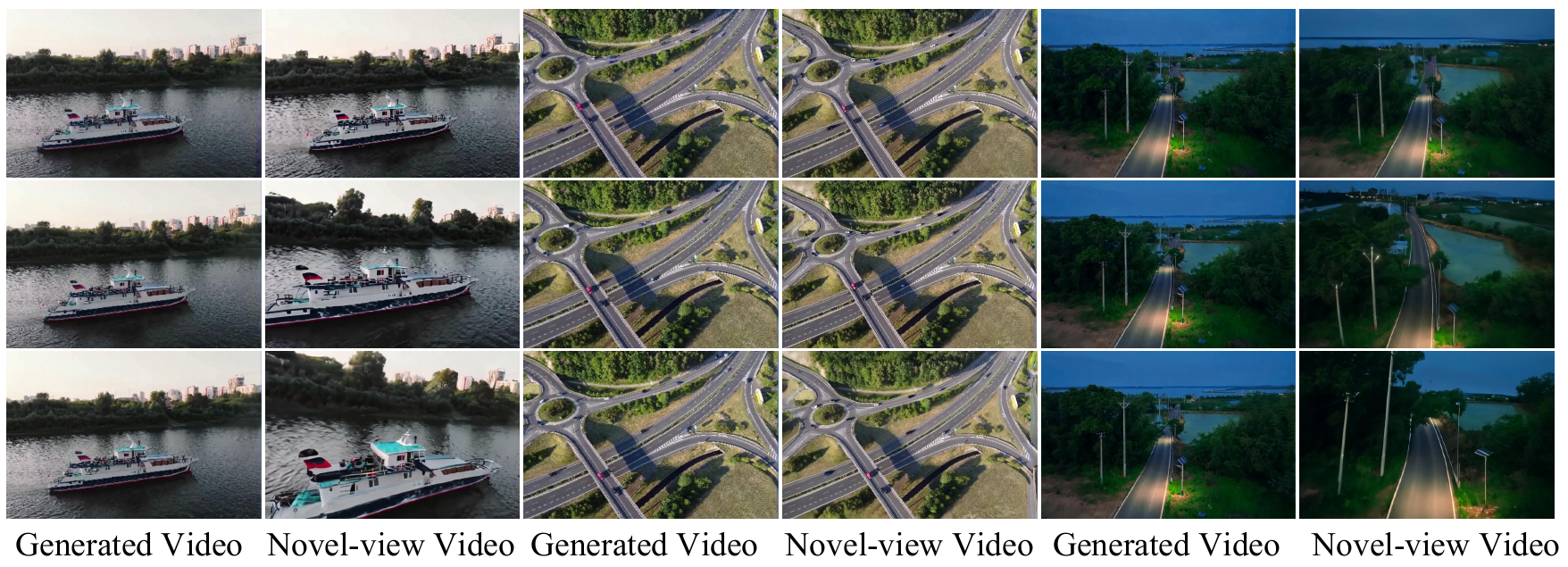}
\caption{\textbf{Novel-view video results on in-the-wild data.}}
\label{fig:4d_result}
\end{figure}

\begin{figure}[t]
\centering
\includegraphics[width=\textwidth]{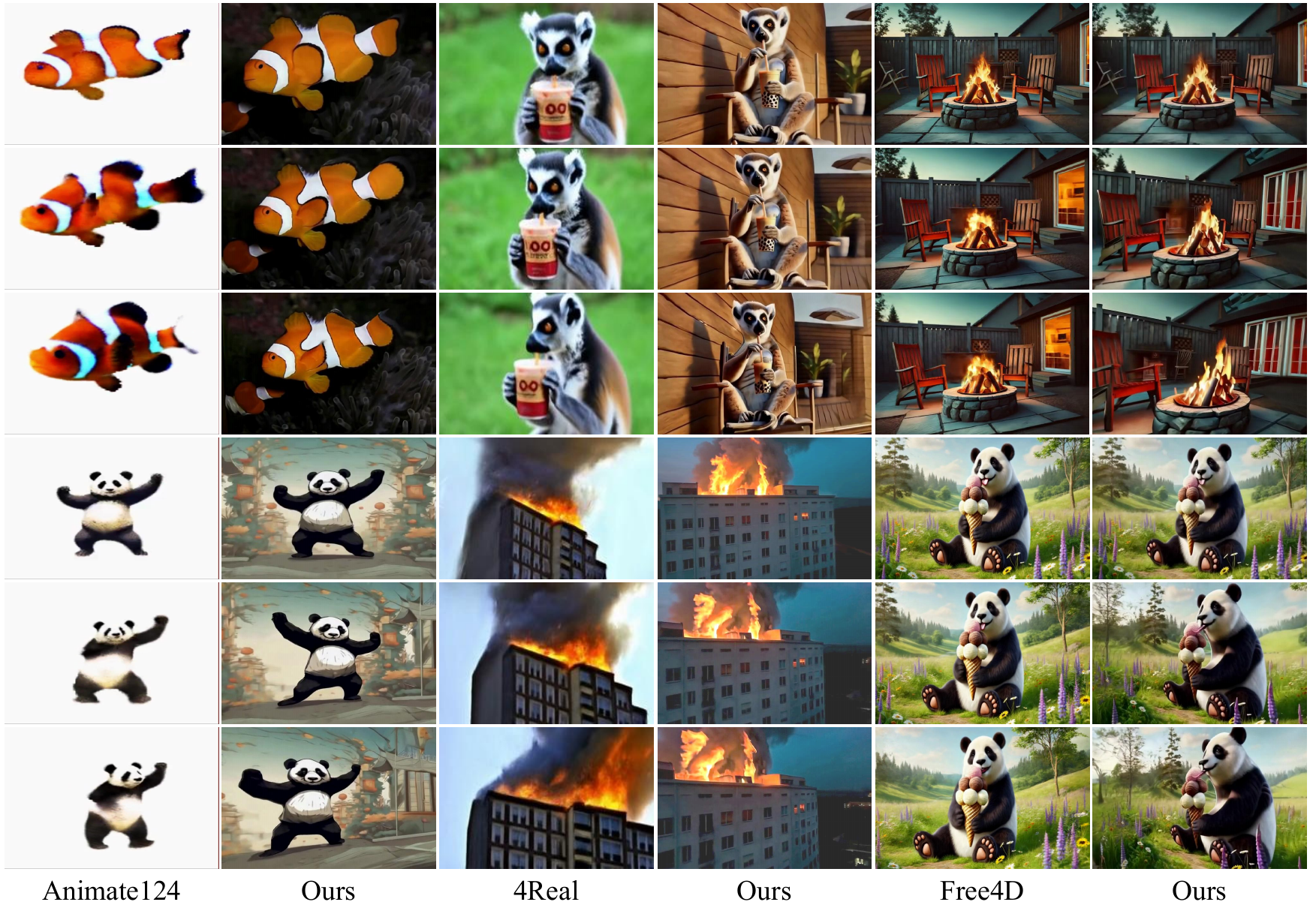}
\caption{\textbf{Qualitative comparison.} Our method generates results with higher consistency, better aesthetics, and notably larger motion than existing 4D generation methods~\cite{animate124,4real,free4d}.}
\label{fig:vs_free4d}
\end{figure}

\begin{figure}[t]
\centering
\includegraphics[width=\textwidth]{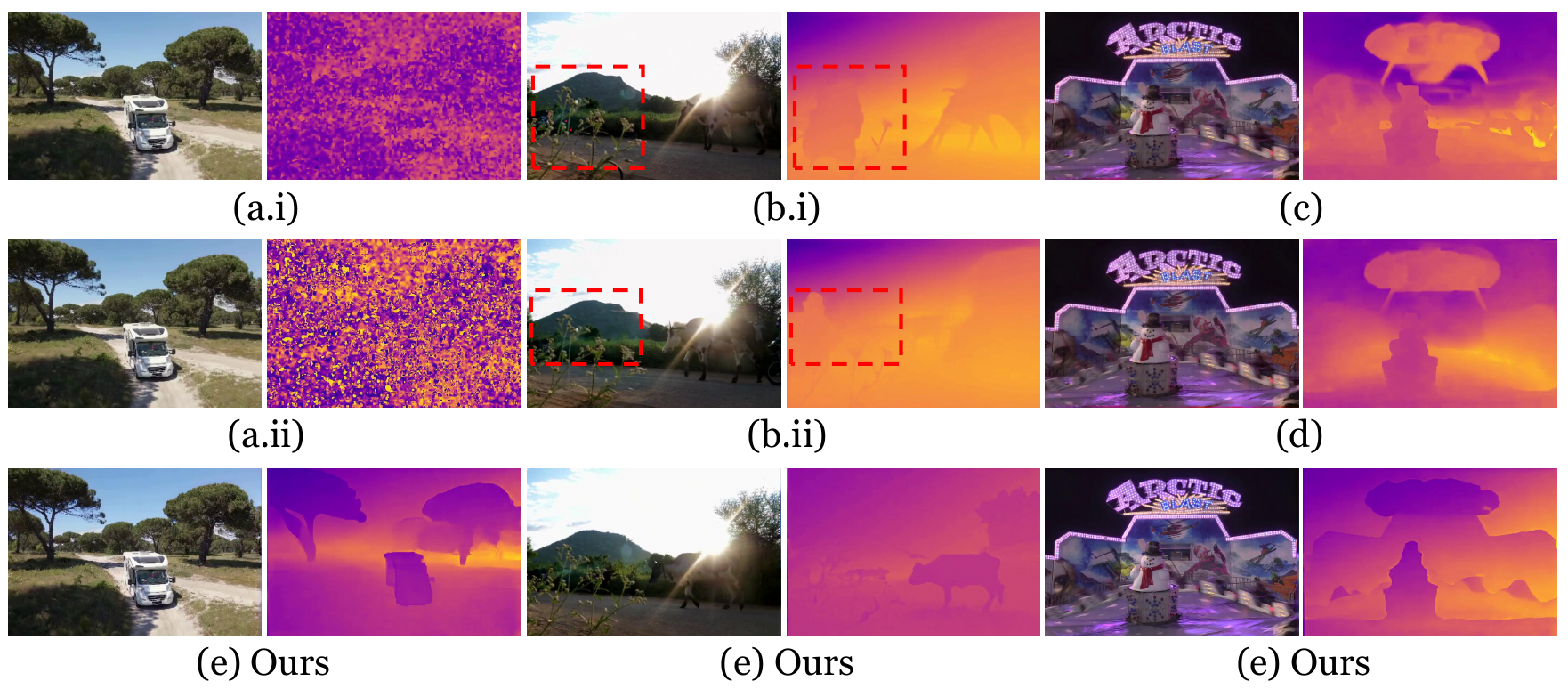}
\caption{\textbf{Ablation study on fusion strategies.} 
We compare channel-wise (a), batch-wise (b), frame-wise (c), height-wise (d), and our width-wise fusion (e) for RGB and XYZ inputs.}
\label{fig:ab}
\end{figure}

\begin{table}[!t]
  \centering
  \setlength{\tabcolsep}{5pt}
  \caption{
    \textbf{4D Generation Results on VBench~\cite{vbench}.} 
    We report the consistency, dynamics, and aesthetics of the generated videos, together with the inference time of each method. 
    }
  \label{tab:image-benchmark-vbench}
  \begin{tabular}{c|cccc}
    \toprule
     Method & Consistency $\uparrow$ & Dynamic $\uparrow$ & Aesthetic $\uparrow$ & Time (min) $\downarrow$\\ \midrule
     4Real~\cite{4real}  
     & 95.7\%    & 32.3\%  & 50.9\%  & 90 \\
     Free4D~\cite{free4d}
     & 96.0\%    & 47.4\%  & 64.7\% & 60 \\
     Ours
     & 96.4\% & 58.0\%  & 59.5\% & 15 \\
     \midrule
     Animate124~\cite{animate124}  
     & 90.7\%    & 45.4\%    & 42.3\%  & \textbackslash \\
     Free4D~\cite{free4d}
     & 96.9\%    & 40.1\%    & 60.5\%  & 60 \\
     Ours
     & 97.2\%  & 58.3\%  & 53.0\% & 15 \\
     \midrule
     % DimensionX~\cite{dimensionX}
     % & \textbf{97.2\%}      & 21.9\%    & 56.0\%  \\
     % Free4D~\cite{free4d}
     % & 95.5\%      & 22.1\%    & \textbf{57.3\%}  \\
     % Ours
     % & 95.5\%    & \textbf{35.4\%}  & 57.2\% \\
     % \midrule
     GenXD~\cite{genXD}
     & 89.8\%    & 98.3\%   & 38.0\% & \textbackslash \\
     Free4D~\cite{free4d}
     & 96.8\%    & 100.0\%  & 57.9\% & 60 \\
     Ours
     & 96.8\%    & 100.0\%  & 52.4\% & 15 \\
     \bottomrule
    \end{tabular}
\end{table}

\begin{table}[!t]
  \centering
  \caption{\textbf{User study results.} Percentages indicate user preference.}
    \begin{tabular}{l c c c}
      \toprule
      Comparison & Consistency & Dynamic & Aesthetic  \\
      \midrule
      Ours vs. Free4D~\cite{free4d}         & 56\% \ / \ 44\% & 59\% \ / \ 41\% & 53\% \ / \ 47\% \\
      Ours vs. 4Real~\cite{4real}           & 79\% \ / \ 21\% & 85\% \ / \ 15\% & 93\% \ / \ 7\% \\
      Ours vs. Animate124~\cite{animate124} & 75\% \ / \ 25\% & 56\% \ / \ 44\% & 100\% \ / \ 0\% \\
      Ours vs. GenXD~\cite{genXD}           & 90\% \ / \ 10\% & 85\% \ / \ 15\% & 100\% \ / \ 0\% \\
      \bottomrule
    \end{tabular}
  \label{tab:user_study}
\end{table}

\subsection{Setting}

\textbf{Baselines.} Following~\cite{free4d}, we compare our method with existing 4D generation approaches, which can be grouped into two categories: text-to-4D and image-to-4D methods. For text-to-4D, we compare against 4Real~\cite{4real}, a state-of-the-art method in this category.  
For image-to-4D, we benchmark against the state-of-the-art Free4D~\cite{free4d}, the feed-forward method GenXD~\cite{genXD}, and the object-levle approach Animate124~\cite{animate124}. For text-to-4D methods, we first generate an image from the input text prompt and then convert it into the image-to-4D setting.
To ensure fairness, we use the same single-image or text prompt across all methods during evaluation.

\textbf{Datasets and Metrics.} 
We conduct evaluations on a collection of images and texts sourced from the official project pages of the compared methods. To assess the quality of generated novel-view videos, We report standard VBench metrics~\cite{vbench}, including Consistency (averaged over subject and background), Dynamic Degree, and Aesthetic Score. Given the lack of a well-established benchmark for 4D generation, we further conduct a user study involving 23 evaluators to enhance the reliability of our evaluation.

\textbf{Implementation Details.} 
We opt for the vanilla Wan2.1~\cite{wan2025wan} image-to-video model as our final base model with a total of 14B parameters\footnote{https://huggingface.co/Wan-AI/Wan2.1-I2V-14B-480P}.
Most importantly, given the significant distribution gap between the spatial coordinates XYZ and the original RGB domain, one may carefully deal with the normalization of the input data to the diffusion model so that the noise scheduling is balanced across two modalities.
Recall our diffusion target is jointly denoising RGB and XYZ where the noised RGB latent is in the space of KL-regularized VAE whose distribution is close to a Gaussian Distribution. 
However, the XYZ coordiante is not normally distributed in the 3D space, which leads to modality gap during denoising.
To bridge this gap, we propose to perform modality-aware normalization. 
Specifically, we trace the statistics (mean and standard deviation) of XYZ domain in the latent space over 5K random samples from the training dataset. 
It results in $\mu = -0.13$ and $\sigma = 1.70$, which serves as the constant normalization term for XYZ latent during training and inference.
To fully transfer the capability of original image-to-video generation from the base model to the target image-to-4D task, we train a LoRA with a rank of 64 for the sake of parameter and data efficiency instead of full-parameter supervised finetuning.
The Lora finetuning is run with a batch size of 32 using an
AdamW optimizer. 
The learning rate is set to $1 \times 10^{-4}$ with a cosine learning rate warmup. 
The training is distributed on 32 NVIDIA A100 GPUs with 5k iterations at a spatial resolution of $480\times 720$ for each modality. To generate novel-view videos, we first produce a 4D point cloud representation of the scene using our feed-forward model, and then render the results using~\cite{trajectorycrafter}.

\subsection{Main Results}

\textbf{4D Geometry Generation.}
As illustrated in Fig.~\ref{fig:geo_result}, we visualize the paired RGB and XYZ video generated from a single image. The results demonstrate that our method can simultaneously infer plausible scene motion and the corresponding 4D geometry from a single image. This high-quality geometric representation of dynamic scenes is essential for consistent and photorealistic novel view synthesis in the subsequent rendering stage.

\textbf{Novel-View Video Generation.}
Quantitative results on VBench~\cite{vbench} are presented in Table~\ref{tab:image-benchmark-vbench}. Our method achieves performance comparable to state-of-the-art approaches, and notably outperforms others in terms of Dynamic Degree.  
Free4D~\cite{free4d} benefits from the proprietary Kling~\cite{kling} model for image animation, which contributes to its higher aesthetic scores.
Qualitative comparisons are shown in Fig.~\ref{fig:vs_free4d}, where our results demonstrate more significant and coherent scene dynamics, especially under camera motion.
Furthermore, user study results (Table~\ref{tab:user_study}) show that our method is consistently preferred over most baselines in terms of consistency, dynamics, and aesthetics.
Although the results are comparable to Free4D, it is important to note that the evaluation was conducted on the Free4D test set, which predominantly features object-centric scenes. In contrast, our method generalizes well to more diverse, in-the-wild scenarios, as illustrated in Fig.~\ref{fig:4d_result}.
In addition, our method is feed-forward and highly efficient, capable of generating a dynamic 4D scene within 15 minutes.
By comparison, Free4D relies on a time-consuming pipeline, typically requiring over one hour to produce results.

\subsection{Ablations and Analysis}
\label{subsec:ab}
To validate the effectiveness of our used width-wise fusion strategy and support the analysis presented in Sec.~\ref{subsec:fusion}, we conduct an ablation study comparing five different fusion designs, as illustrated in Fig.\ref{fig:ab}.
Among these, channel-wise fusion introduces a severe distribution mismatch with the pretrained prior, often leading to noisy or failed predictions (\textit{a.i-a.ii}). Batch-wise fusion preserves unimodal quality but fails to capture cross-modal alignment, yielding inconsistent RGB-XYZ correlation (\textit{b.i-b.ii}). Frame-wise (\textit{c}) and height-wise (\textit{d)} strategies provide moderate improvements, yet still suffer from suboptimal alignment and visual quality.
In contrast, our width-wise fusion brings corresponding RGB and XYZ tokens closer in the sequence, significantly shortening the cross-modal interaction distance. This facilitates more effective alignment and yields sharper, more consistent geometry and appearance across frames, as demonstrated in Fig.~\ref{fig:ab}~(e).
\section{Conclusion}
\label{sec:conclusion}

We present \methodname, the first feed-forward framework for generating 4D scene representations from a single image. Our approach fine-tunes a pretrained video diffusion model to enable efficient image-to-4D generation. 
To address the scarcity of training data, we construct \datasetname, a large-scale dataset with high-quality pseudo-4D annotations. Furthermore, we propose a unified 6D video representation that jointly models appearance and geometry, along with a set of simple yet effective adaptation strategies to repurpose video diffusion models for the 4D generation task. 
Extensive experiments demonstrate \methodname generates high-quality dynamic point clouds, providing a reliable geometric foundation for synthesizing novel-view videos. The resulting videos achieve competitive performance compared to existing methods, while offering superior efficiency and generalizability. 
% We hope this work paves the way for scalable and accessible single-image generative 4D world modeling.

\paragraph{Limitations and Future Work}
While \methodname demonstrates promising results in single-image 4D generation, several limitations remain. 
First, our method relies on pseudo-4D annotations for supervision, which may introduce noise or inconsistencies, particularly in fine-grained geometry or long-term temporal coherence. 
Introducing high-quality real-world or synthetic dataset would be fruitful for general 4D modeling.
Second, although the image-driven generated results are 4D-grounded, controllabilities over lighting, fine-grained motion and physical property are still lacking.
Third, the unified 6D representation, while effective, assumes relatively clean input images and may degrade under occlusions, extreme lighting conditions, or cluttered backgrounds. 
Future work includes improving temporal modeling with explicit world priors, incorporating real-world 4D ground-truth data when available, and extending our framework to handle multi-object or interactive scenes. 
Additionally, integrating multi-modal inputs like text or audio could further enhance controllability and scene diversity.
\bibliography{iclr2026_conference}
\bibliographystyle{iclr2026_conference}

\clearpage
\appendix
\setcounter{figure}{0}
\setcounter{section}{0}
\setcounter{table}{0}
\renewcommand{\thefigure}{{\Alph{figure}}}
\renewcommand{\thetable}{{\Alph{table}}}
\renewcommand{\thesection}{{\Alph{section}}}

\section*{Appendix}
\addcontentsline{toc}{section}{Appendix}

% This supplementary material is organized as follows:
% \begin{itemize}[leftmargin=*, itemsep=2pt, topsep=2pt]
%     \item Section.~A provides details of our user study.
%     \item Section.~B presents additional implementation details about ablation studies.
% \end{itemize}

\section{Details of User Study}
\label{supp:details-of-user-study}

\begin{figure}[t]
\centering
\includegraphics[width=0.75\textwidth]{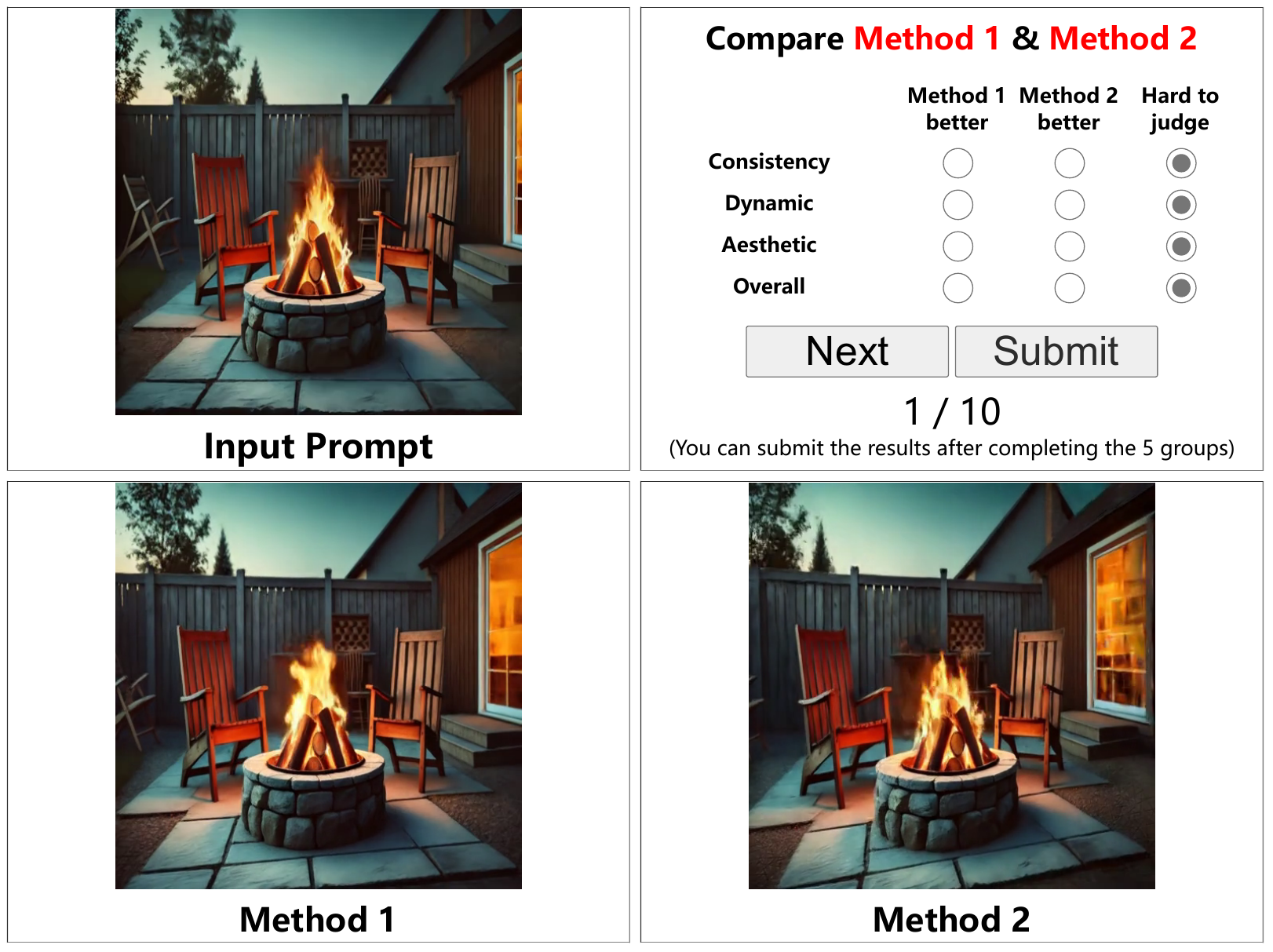}
\caption{
\textbf{User study interface.} Participants were shown an input prompt and two generated videos from different methods. They were asked to compare the results based on \textit{Consistency}, \textit{Dynamics} and \textit{Aesthetics}. Each question allowed skipping if the difference was hard to judge.
}
\label{fig:userstudy-ui}
\end{figure}

\noindent \textbf{User Study: Comparison with Existing Methods.}  
To evaluate the effectiveness of our method, we conducted a user study comparing it against several existing approaches. We collected a total of $74$ video pairs, each generated from the same input image or text prompt to ensure fair comparisons. Competing methods included Free4D~\cite{free4d}, 4Real~\cite{4real}, GenXD~\cite{genXD}, and Animate124~\cite{animate124}. All comparison videos were obtained from their official project pages.
The study was conducted online, and a screenshot of the evaluation interface is shown in Fig.~\ref{fig:userstudy-ui}. Participants were asked to assess each video pair across three criteria: Consistency, Dynamics, and Aesthetics. For each criterion, they were instructed to choose the video they perceived as better. If a comparison was too difficult to judge, they could skip to the next example without selecting an answer.
All responses were collected anonymously, and no personal data were recorded during the study.

\section{Details of VBench Metrics}
\label{supp:details-of-vbench-metrics}
To comprehensively evaluate the quality of our synthesized novel-view videos, we adopt a suite of metrics introduced in VBench~\cite{vbench}, covering three key aspects: \textit{Consistency} (for both subject and background), \textit{Degree of Motion}, and \textit{Aesthetic Quality}.

\noindent \textbf{Subject / Background Consistency.}  
This metric assesses how consistently both the main subject (\eg, human, vehicle, animal) and the surrounding background are maintained throughout the video. It leverages feature similarity across frames using DINO~\cite{dino} for the foreground and CLIP~\cite{clip} for the background. DINO focuses on preserving subject identity by comparing learned visual representations, while CLIP captures broader scene coherence. The average of both provides a balanced view of overall temporal consistency.

\noindent \textbf{Degree of Motion.}  
To avoid favoring overly static videos that may perform well on consistency metrics, we include a motion-aware measure. Specifically, RAFT~\cite{raft} is applied to estimate optical flow, and the \textit{Dynamic Degree} is computed by averaging the top $5\%$ of largest flow magnitudes. This helps emphasize prominent movements, such as object actions or camera shifts, while de-emphasizing negligible or noisy motions, ensuring a more meaningful evaluation of dynamics.

\noindent \textbf{Aesthetic Quality.}  
To reflect the perceived visual appeal of the generated videos, we utilize the LAION Aesthetic Predictor~\cite{LAION}, a lightweight regressor trained atop CLIP features to score image aesthetics on a scale from $1$ to $10$. It considers multiple factors, including color composition, realism, layout, and overall artistic impression. We apply this predictor to each frame and report the average score as the final \textit{Aesthetic Quality} metric.

\subsection{Cross-Domain Self-Attention (CDSA)}
\label{app:cdsa}

\begin{figure}[t]
\centering
\includegraphics[width=\textwidth]{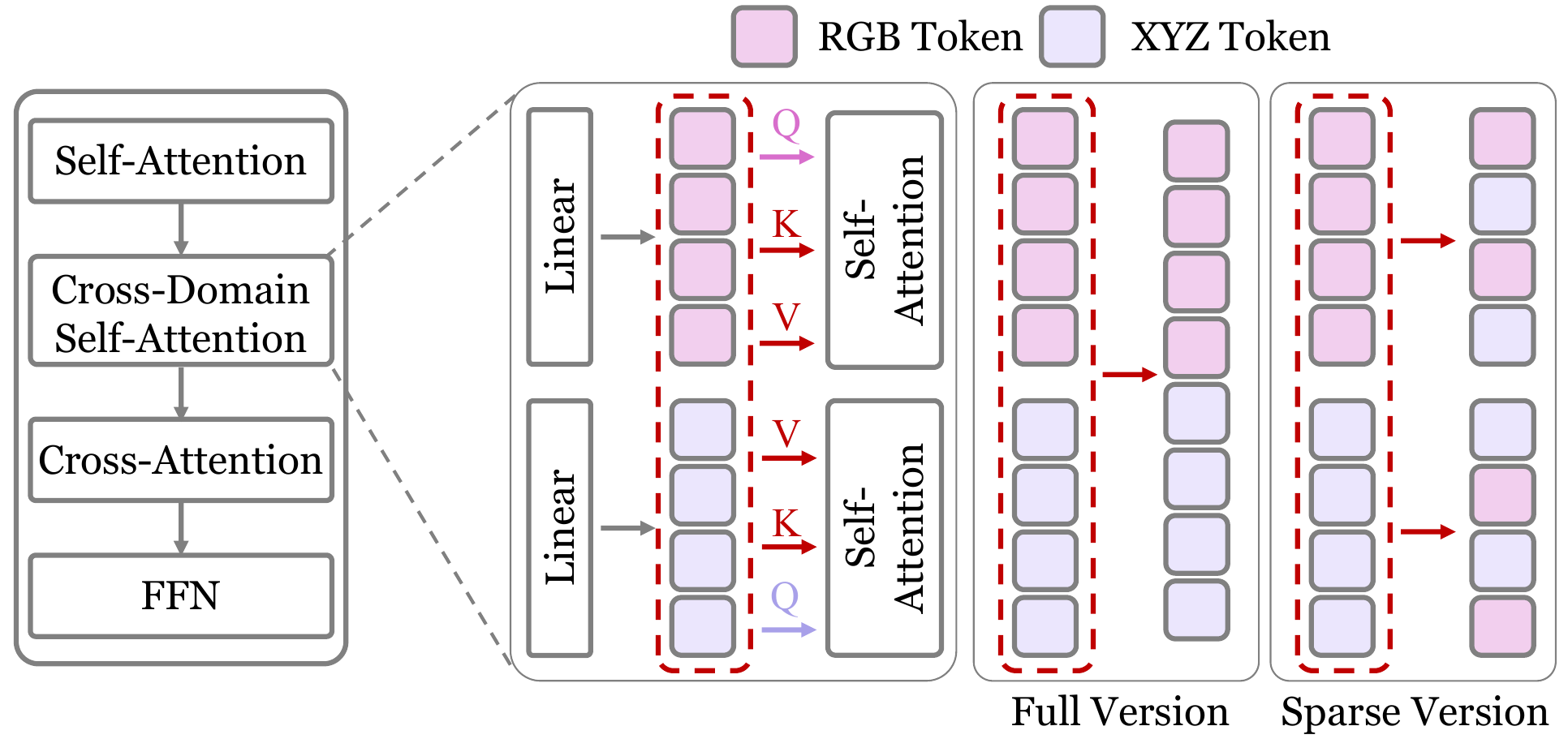}
\caption{
\textbf{Architecture of the Cross-Domain Self-Attention (CDSA) module.} 
The CDSA block is inserted between self-attention and cross-attention layers to facilitate bidirectional interaction between RGB and XYZ modalities. We explore two variants: the \textit{Full Version}, where all tokens interact densely, and the \textit{Sparse Version}, where attention is restricted to spatially corresponding token pairs. This design enables effective cross-modal alignment with different trade-offs in efficiency and performance.
}
\label{fig:cdsa}
\end{figure}

As introduced in Sec.~\ref{subsec:fusion}, we introduce a Cross-Domain Self-Attention (CDSA) module to enhance the alignment between RGB and XYZ modalities, particularly under the batch-wise fusion strategy. Figure~\ref{fig:cdsa} illustrates the architecture of this module.

As shown in the left part of Fig.~\ref{fig:cdsa}, the CDSA block is inserted between the standard self-attention and cross-attention layers within a transformer block. It explicitly enables bidirectional interaction between RGB and XYZ tokens through attention mechanisms—allowing RGB tokens to attend to XYZ tokens and vice versa—thus facilitating cross-modal information exchange.

To balance performance and efficiency, we implement and compare two versions of CDSA:
\begin{itemize}
    \item Full Version: All RGB and XYZ tokens participate in dense cross-domain attention. This version achieves stronger modality interaction at the cost of higher memory and computation.
    \item Sparse Version: Token interactions are restricted to spatially corresponding positions between RGB and XYZ sequences. This reduces overhead while retaining most of the alignment benefits.
\end{itemize}

While both versions aim to bridge the modality gap by promoting fine-grained token-level communication, our experiments reveal that under the batch-wise fusion setting (Fig.~\ref{fig:fusion}~(b.ii)), even with CDSA, the overall cross-modal alignment remains limited. This is primarily due to the spatial separation of RGB and XYZ tokens, which contrasts with the more effective width-wise fusion strategy (Fig.~\ref{fig:fusion}~(e)) where the interaction distance is inherently shorter.

\end{document}